# Canonical Trends: Detecting Trend Setters in Web Data

**Felix Bießmann**  FELIX.BIESSMANN@TU-BERLIN.DE
Dept. ML, Berlin Institute of Technology

**Jens-Michalis Papaioannou**  JENSMICHA@GOOGLEMAIL.COM
Dept. ML, Berlin Institute of Technology

**Mikio Braun**  MIKIO.BRAUN@TU-BERLIN.DE
Dept. ML, Berlin Institute of Technology

**Andreas Harth**  HARTH@KIT.EDU
Institute AIFB, Karlsruhe Institute of Technology (KIT)

## Abstract

Much information available on the web is copied, reused or rephrased. The phenomenon that multiple web sources pick up certain information is often called trend. A central problem in the context of web data mining is to detect those web sources that are first to publish information which will give rise to a trend. We present a simple and efficient method for finding trends dominating a pool of web sources and identifying those web sources that publish the information relevant to a trend before others. We validate our approach on real data collected from influential technology news feeds.

## 1. Introduction

Temporal information is a fundamental aspect of many datasets. Several commercial offerings are based on temporal variation of web sources for data mining[1]. The news domain is a prime example for an industry where time matters. Sources that break a story gain reputation and economic benefits. We thus consider the problem of identifying trendsetting news sources based on temporal correlations found in data.

Our definition of a trend setter is simple. If a single web source publishes content that will later on dominate the content of a pool of other websites, we consider this source as a trend setter; this approach is similar to causality based graph analyses such as in (Lozano and Sindhwani., 2010). In order to test the trendsetting behavior of a web source we first extract a time series of features from each web source. Then we learn for each web source a convolution in this feature space that predicts the content of all other web sources of interest.

Our contributions are as follows:

- We present an approach that detects the *canonical trends* (CTs) in a pool of web sources. The canonical trends capture the information cascades with the highest impact on that collection of web sources.

- We propose an unsupervised algorithm that identifies web sources which predict these trends before they arise. The features necessary to predict the trends are automatically learned; they help to identify information cascades and their temporal dynamics.

- We evaluate the approach on a dataset of news items from 96 popular technology newsfeeds, collected over several months, showing that our approach predicts the temporal evolution of news items better than classical topic detection.

As an example data set we collected data from the most influential technology news websites. Bag-of-Word (BoW) features were extracted for each website.

---

[1] See e.g. http://www.google.com/trends/correlate/

We then predict the total information of all websites at time point $t$ using only the information of one web source at prior time points $t - \tau$. Our results show that some news sites can predict the future temporal dynamics of the tech-news-sphere well, while others fail to do so. The prediction performance can be interpreted as how much a given news site can be considered as a trend setter and can be used to rank sites according to this criterion: The better a news site predicts the future information of all other news sites, the more influential the news site is.

## 2. Related Work

In the following we discuss some alternative approaches towards analysis of temporal dynamics in web data graphs. The authors of (Sun et al., 2007) use the temporal dynamics within a communication network graph to partition the nodes of the graph into groups. The method first extracts adjacency matrices of the graph for different time points and then tries to compress this time series of connections. This is done by finding similar connection patterns over time and group them together. The motivation of this approach is very different from ours and a direct comparison of these two approaches is not possible. But there is a similarity that is worth noting: If one web source predicts the content of all other nodes perfectly, we can focus on this single node only and forget about the rest of the network. Thus the representation found by our approach can be seen as an optimal compression of the graph, too.

Other approaches towards network data graphs evolving over time investigate the diffusion of influential items, so called *memes* (Leskovec et al., 2009; Yang & Leskovec, 2010; Gomez Rodriguez et al., 2011). In (Leskovec et al., 2009; Yang & Leskovec, 2010) the authors focus on diffusion of n-grams in blogs and news media. The method proposed in (Yang & Leskovec, 2010) finds those n-grams that are repeated often, i.e. that account for a large volume of a graph. This objective is very similar to that of this study. The objective of our method is to predict the content of a pool of web sources optimally. This is equivalent to finding nodes that maximize the variance explained of a pool of other web sources. Similar to (Yang & Leskovec, 2010) we use a linear model. A decisive advantage of our approach is that it straightforwardly extends to non-linear dependencies (see section 5.2). Another important difference is that in (Yang & Leskovec, 2010) information transmission is modeled as an indicator function in, meaning information has been transmitted at a certain time lag or not. In our approach we do not restrict the analysis to a binary transmission scheme. Instead we learn a gradual information transmission model from the data. Another related approach is (Gomez Rodriguez et al., 2011). Here the authors analyze the temporal dynamics of information cascades in a temporally evolving graph, in particular how n-grams diffuse through a network. The cascades are represented as time stamps of selected n-grams. Different generative models are fitted to the data using convex optimization. A central assumption is that the transmission rates can be estimated independently for each cascade. This assumption is similar to our approach: We analyze the temporal dynamics of single web sources independently.

Despite a number of similarities between (Leskovec et al., 2009; Yang & Leskovec, 2010; Gomez Rodriguez et al., 2011) and our method we emphasize an important difference: All of the above approaches require that the relevant items of information are selected prior to the analysis. For example in (Leskovec et al., 2009; Yang & Leskovec, 2010) the authors analyze a large data set containing millions of n-grams. But only 1000 information cascades are selected for the final analysis according to some heuristics. Thus the result can depend on data selection during preprocessing. Our approach is different in that it takes the full data set and automatically learns the relevant features. Another crucial difference is that the above approaches do not model dependencies between information cascades. In real data sets it is very likely that on piece of information is highly correlated with another piece of information. The method proposed here takes into account the dependencies between features and models the full multivariate temporal dynamics between web sources.

## 3. Canonical Trends

For our approach we extract from each web source $f \in \{1, 2, \ldots, F\}$ in our collection of $F$ web sources the corresponding features $x_f(t) \in \mathbb{R}^W$ at time points $t = \{0, 1, \ldots, T\}$. For the sake of simplicity we here assume regularly sampled time points. In our application example we will extract Bag-of-Words features, see section 6.2.1, but our approach is readily applicable to other feature representations such as n-grams or collections of hyperlinks. After feature extraction we store the multivariate feature time series in a sparse matrix

$$X_f = [x_f(t=1), \ldots, x_f(t=T)] \in \mathbb{R}^{W \times T}. \quad (1)$$

We are interested not in the dynamics of a single web source but rather the temporal variation of many nodes in the web graph. The joint time series of all web sources $Y_f$ can be obtained as the average across all $X_f$

$$Y_f = 1/(F-1) \sum_{f' \neq f} X_{f'} \in \mathbb{R}^{W \times T} \quad (2)$$

where $f'$ denotes the indices of all web sources except $f$. We now represent a canonical trend (CT) $y_f(t)$ as a combination $w_y \in \mathbb{R}^W$ of features

$$y_f(t) = w_y^\top Y_f(:,t) \quad (3)$$

In our application example of BoW features we chose a linear feature combination as the optimal tradeoff between a too simplistic modeling of single word occurrences[2] and a computationally costly n-gram representation as e.g. in (Leskovec et al., 2009). If the relationships between single features are more complex, the linear feature combination $w_y$ can be replaced by arbitrary non-linear feature combinations simply by using appropriate kernel functions (see section 5.2).

## 4. Canonical Trend Prediction

The aim of our approach is to predict the temporal variation of the overall trend $y_f(t)$ using the information published in the past $N_\tau$ hours by a single news feed $X_f$. This means we want to find a temporal convolution $w_x(\tau)$ that uses the information of $x(t-\tau), \tau \in \{1, \ldots, N_\tau\}$ to predict the canonical trend $y_f(t)$. The optimal prediction of $y_f(t)$ based on the content published in the past $N_\tau$ hours on a single web source $x_f(t)$ can be formulated as

$$\hat{y}_f(t) = \sum_\tau w_x(\tau)^\top X_f(:, t-\tau). \quad (4)$$

Neglecting the amplitude of $y_f(t)$ and $\hat{y}_f(t)$, minimizing the least-squares error of eq. 4 is equivalent to maximizing the correlation between $y_f(t)$ and $\hat{y}_f(t)$

$$\underset{w_x(\tau), w_y}{\operatorname{argmax}} \operatorname{Corr}(y_f(t), \hat{y}_f(t)). \quad (5)$$

The optimal $w_x(\tau)$ and $w_y$ can be computed simultaneously using canonical correlation analysis (CCA) (Hotelling, 1936). CCA has proven very useful for a wide variety of applications ranging from signal processing (Akaike, 1976) over efficient computation of causality measures (Otter, 1991). The mathematical properties of CCA are as well understood (Jordan, 1875) as its statistical convergence criteria (Anderson, 1999; Fukumizu et al., 2007). Instead of standard CCA we use an extension, temporal kernel CCA (tkCCA), that can deal with high dimensional data, small sample sizes and time delayed non-linear dependencies between data (Bießmann et al., 2010). The interpretation of $w_y$ and $w_x(\tau)$ is straightforward. In our application example they are the directions in the BoW feature space that maximize the correlation between a single feed and all other news feeds (or equivalently – assuming normalized time series – minimize the prediction error between the two). CCA simultaneously optimizes $w_y$ and $w_x(\tau)$ such that the correlation between $y_f(t)$ and $\hat{y}_f(t)$ is invariant with respect to all linear transformations of the data[3]. This is why the correlation coefficient in eq. 5 is called *canonical*. The projection $w_y$ maps the data into their respective *canonical subspace*. We thus refer to the time series $y_f(t)$ as the *canonical trend* (CT) in the BoW feature space.

---

[2]See e.g. http://www.google.com/trends/correlate/.

The correlation coefficient in eq. 5 is obtained from a convolved time series. The convolution in eq. 4 sums over all time lags $\tau$. Often it can give valuable insights in the temporal dynamics between variables if one computes a time lag dependent correlation coefficient $\rho(\tau)$

$$\rho(\tau) = \operatorname{Corr}(w_x(\tau)^\top X_f(:, t-\tau), w_y^\top Y_f(:, t)). \quad (6)$$

We will refer to $\rho(\tau)$ as the *canonical correlogram*, in complete analogy to a standard cross-correlogram. The main difference is that standard cross-correlograms are typically computed between two univariate signals. The *canonical correlogram* is computed between high dimensional multivariate time series, projected into their canonical subspace. The canonical correlogram $\rho(\tau)$ and the coefficients of the convolution $w_x(\tau)$ reflect the temporal dynamics in the canonical subspace. An illustrative toy data example is shown in figure 1, for an explanation see section 6.1.

## 5. Canonical Trend Algorithm

Informally our approach consists of three steps:

1. Extract feature matrix $X_f$ for each feed

2. Temporal Embedding of single news feed $X_f$

3. Kernel CCA between $X_f$ and all other feeds $Y_f$

In the following we describe steps two and three in detail. Data collection and feature extraction are described in section 6.2.1.

---

[3]Or invariant w.r.t. non-linear transformations in the case of kernel CCA

## 5.1. Temporal Embedding

The temporal embedding is done by creating for each feed $f$ a new representation $\tilde{X}_f$ in which we add copies of the data in $X_f$, shifted back in time by a time lag $\tau$

$$\tilde{X}_f = \begin{bmatrix} X_{f,\tau=-N_\tau} \\ \vdots \\ X_{f,\tau=-1} \end{bmatrix} \in \mathbb{R}^{WN_\tau \times T}. \qquad (7)$$

## 5.2. Kernel CCA

The temporal embedding operation will increase the dimensionality of our data by a factor of $N_\tau$, the number of time lags. However using the well known kernel trick (Aizerman et al., 1964) we can efficiently compute CCA in kernel space. A main advantage of this trick is that computation of non-linear dependencies becomes a linear problem in kernel space, see e.g. (Fyfe & Lai, 2000). Another crucial advantage of kCCA for the given problem setting is that it reduces the problem size substantially: Estimating $w_y$ and $w_x(\tau)$ in the input space requires the inversion of covariance matrices of size $(W + WN_\tau)^2$, where $W$ denotes the number of features. In kernel space we only have to deal with matrices of size $(2T)^2$, where $T$ denotes the number of samples. For the sake of simplicity we consider linear kernels here, but non-linear dependencies can be easily estimated by replacing the linear kernel with other kernel functions. When using linear kernels the CCA solution in input space is a linear expansion of data points

$$w_x(\tau) = X_{f,\tau}\alpha, \qquad (8)$$
$$w_y = Y_f\beta. \qquad (9)$$

The coefficients $\alpha$ and $\beta$ the eigenvectors of the generalized eigenvalue problem

$$\begin{bmatrix} 0 & K_xK_y \\ K_yK_x & 0 \end{bmatrix} \begin{bmatrix} \alpha \\ \beta \end{bmatrix} = \lambda \begin{bmatrix} L_x & 0 \\ 0 & L_y \end{bmatrix} \begin{bmatrix} \alpha \\ \beta \end{bmatrix} \qquad (10)$$

where $K_x = \tilde{X}_f^\top \tilde{X}_f \in \mathbb{R}^{T \times T}$ is the linear kernel matrix of $\tilde{X}_f$ and $K_y = Y_f^\top Y_f \in \mathbb{R}^{T \times T}$ is the linear kernel matrix of $Y_f$. The eigenvalue $\lambda$ is the canonical correlation on the training data set[4], which yields the same result as eq. 5. The matrices on the right hand side are computed as $L_x = K_x^2 + \kappa I$ and $L_y = K_y^2 + \kappa I$, where $\kappa$ is a regularization parameter controlling the complexity of the solution. For very noisy data $\kappa$ will

---

[4]We consider here only the first dimension of the canonical subspace corresponding to the first eigenvalue; multidimensional canonical subspaces can be found by solving eq. 10 for more than one eigenvalue.

---

**Algorithm 1** Canonical Trend Algorithm

**Input:** Data $\{X_{f=1}, \ldots, X_{f=F}\} \in \mathbb{R}^{W \times T}$,
  optimal time lag $N_\tau$, optimal regularizer $\kappa$
Loop over all news feeds
**for** $f = 1$ **to** $F$ **do**
  Average over all news feeds except $f$
  $Y_f = 1/(1-F)\sum_{f' \neq f} X_{f'}$
  Temporal Embedding (eq. 7)
  $\tilde{X}_f = [X_f(:, t-N_\tau)^\top, \ldots, X_f(:, t-1)^\top]^\top$
  Compute Kernels
  $K_x = \tilde{X}_f^\top \tilde{X}_f$
  $K_y = Y_f^\top Y_f$
  Cross-validation loop
  **for** fold = 1 **to** 10 **do**
    Pick Training indices Tr $\in \{1, \ldots, T\}$
    Pick Test indices Te $\in \{1, \ldots, T\} \setminus$ Tr$-N_\tau$
    $\alpha, \beta = \text{kCCA}(K_x(\text{Tr}, \text{Tr}), K_y(\text{Tr}, \text{Tr}), \kappa)$
    Predict Test data
    $c_{f,\text{fold}} = \beta^\top K_y(\text{Tr}, \text{Te}) K_x(\text{Te}, \text{Tr}) \alpha$
  **end for**
**end for**
Rank Feeds according to $1/10 \sum_{\text{fold}} c_{f,\text{fold}}$

---

be large and thus the optimal $\alpha$ and $\beta$ will be a vector with very similar entries $\alpha_i$, and the same will be true for $\beta$. This effectively means that eq. 8 and eq. 9 will reduce to computing the empirical mean of the data. After optimization of $N_\tau$, $\kappa$ and eq. 10 we can recover the canonical projection $w_y$ according to eq. 8 and the canonical convolution $w_x(\tau)$ according to eq. 9. We then could compute $\hat{y}_f(t)$ according to eq. 4 and the overall trend $y_f(t)$ using eq. 3. In practice however this is suboptimal in terms of computational cost. Instead of recovering $w_y$, $w_x(\tau)$ and computing $y_f(t)$, $\hat{y}_f(t)$, we can stay in kernel space to evaluate the models. This yields a substantial computational speedup once the kernels are computed. The complete canonical trend detection algorithm is summarized in algorithm 1.

## 5.3. Model evaluation for time series

In order to obtain meaningful prediction accuracies we apply 10-fold cross-validation: we split the available data into training and test data, estimate $\alpha$ and $\beta$ on the training set and compute the prediction accuracy in eq. 5 on test data. When performing cross-validation on time series data special care has to be taken. In contrast to standard classification settings, where one can simply randomly pick a certain subset of the data, the temporal dependencies in time series data do not allow for such a simple resampling. For proper cross-validation we split the time series in 10 blocks of equal length. Due to the temporal embedding (see eq. 7) consecutive blocks will overlap by $N_\tau$ samples. Thus we discarded the first $N_\tau$ samples from the training block adjacent to the test data block. This ensured that no data point that we tested on was used for training the KCCA model. We estimated the optimal time lag and regularization parameters using 10-fold cross-validation (nested within the training data set) and a grid search over time lags $\tau \in \{1, 2, \ldots, 10\}$ and $\kappa \in \{10^{-5}, 10^{-4}, \ldots, 10^1\}$. Optimal regularizers $\kappa$ were in the range of $10^{-3}$ to $10^{-1}$, the optimal time lag was $\tau = 5$ hours.

### 5.4. Comparison with other approaches

The relevant contribution of the CT algorithm is that it maximizes the *co-variation* of single web sources $X_f$ and other web sources $Y_f$. This is accomplished by a joint factorization of $\tilde{X}_f$ and $Y_f$ (see eq. 10). An alternative approach for topic detection is *latent semantic analysis* (LSA) (Deerwester et al., 1990) in which only a single matrix of BoW features is factorized. In LSA the strongest topic $v_{y,f} \in \mathbb{R}^W$ is that subspace in the BoW space, here the row space of $Y_f$, that captures most variance

$$\underset{v_{y,f}}{\operatorname{argmax}}(v_{y,f}^\top Y_f Y_f^\top v_{y,f}), \qquad \text{s.t. } v_{y,f}^\top v_{y,f} = 1. \quad (11)$$

The strongest topic $v_{x,f}$ in the single feed BoW space $X_f$ is found analogously. Informally the relationship between LSA and CT is similar to the relationship between principal component analysis (PCA) (Pearson, 1901) and CCA: PCA maximizes the variance *within* one web source $X_f$ (or a collection of web sources $Y_f$) while CCA maximizes the *co-variation between* multiple web sources $X_f$ and $Y_f$. We compared the canonical trend predictions (eq. 5) with the correlation between $v_{y,f}^\top Y_f$ and $v_{x,f}^\top X_f$ obtained by LSA on $X_f$ and $Y_f$ separately. As an additional sanity check we also shuffled the data in time and thereby destroyed the temporal dependencies between $X_f$ and $Y_f$ (results shown in table 1, middle column). All analyses were performed analogously on this surrogate data set, in order to show that the prediction accuracies were indeed meaningful and not just overfitted.

## 6. Results

We first illustrate our approach on a toy data set. Thereafter we present some results on real data extracted from technology news feeds.

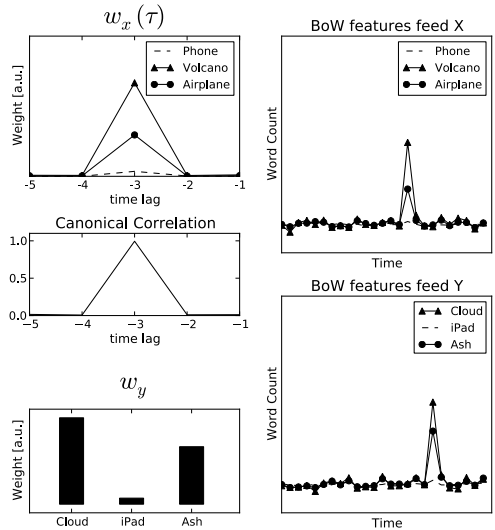

*Figure 1.* A toy data example (see section 6.1). *Right Panels:* News feed X reports on the eruption of Eyjafjallajoekull 3 hours before news feed Y. *Left Panels:* Solution of Canonical Trend Detection; $w_y$ (*bottom left*) has high coefficients for 'Cloud' and 'Ash'; at a time lag of $\tau = -3$ $w_x(\tau)$ (*top left*) has high coefficients for 'Volcano' and 'Airplane'. Irrelevant words have low weights. Temporal dynamics are also captured by the correlogram $\rho(\tau)$ (*middle left panel*) in the canonical subspace (see eq. 6): A peak at $\tau = -3$ indicates that news feed $X$ is 3 time samples ahead of $Y$.

### 6.1. Canonical Trends: A toy data example

For illustrative purposes we consider an event that has been reported extensively on. In 2010 a volcano on Iceland erupted and produced a large ash cloud. Due to this cloud a lot of flights had to be cancelled for security reasons, as the ash could damage aircraft turbines. In the course of the events, every news page on the web reported on the eruption and its consequence. Not every news page used the same words but the overall trend across all news pages included words like *eruption, volcano, iceland, aircraft, traffic* etc. that co-occurred increasingly. We model this trend in the BoW feature representation time series of two different web sources $X \in \mathbb{R}^3$ and $Y \in \mathbb{R}^3$. The trend is reflected in the different dimensions of $X$ with a weighting $w_x^* = [0.05, 0.9, 0.4]^\top$ corresponding to the words *Phone, Volcano, Airplane* and analogously it is reflected in the dimensions of $Y$ with the weighting $w_y^* = [0.9, 0.05, 0.6]^\top$ corresponding to the words *Cloud, iPad, Ash*. So one BoW dimension did not carry relevant information (*Phone, iPad*) and the other two dimensions did carry relevant information, respectively. The toy data was generated from an underlying trend $s(t) \in \mathbb{R}^1$, reflecting the volcano eruption and its consequences on air traffic, by

$$X(:,t) = \gamma w_x^* s(t-3) + \sqrt{1-\gamma}\epsilon_x(t) \qquad (12)$$
$$Y(:,t) = \gamma w_y^* s(t) + \sqrt{1-\gamma}\epsilon_y(t)$$

where $\epsilon(t) \sim \mathcal{N}(0,1)$ was noise drawn from a standard normal distribution in $\mathbb{R}^3$ and $\gamma = 0.9$ was the signal to noise ratio of the trend. The BoW time series are shown in the right panels of figure 1. $X$ is generated from the latent trend variable $s(t)$ with a temporal lag of $-3$ temporal units so that $X$ will be ahead of $Y$ by 3 time samples. Note that the dimensions in $X$ were not related to the dimensions of $Y$. This is a realistic setting: In practice this is difficult to define all possible trends *a priori*, even with the help of a semantic dictionary. But the increased co-occurrence of the above mentioned trend-relevant words, that is the temporal co-variation in the *canonical subspace* defined by $w_x^*$ and $w_y^*$, captures the trend very well.

This canonical subspace is robustly found by the canonical trend detection algorithm. The optimal convolution $w_x(\tau)$ and the projection $w_y$ are plotted on the left of figure 1. They clearly reflect the structure of $w_x^*$ and $w_y^*$ that gave rise to the trend in the BoW space. In the case of $w_x^*$, the canonical trend detection yields a convolution, rather than a simple projection. The additional temporal dimension indicates the temporal dependency structure in the canonical subspace. At a time lag of $-3$ temporal units, the web source $X$ predicts the web source $Y$ best. So the optimal BoW features for $X$, corresponding to $w_x^*$, are found at a time lag of $-3$. The canonical correlogram (see eq. 6) for our toy data example is plotted in the middle panel on the left and shows a strong peak at $\tau = -3$, indicating that $X$ published the relevant information 3 time units before $Y$.

### 6.2. Trend setter detection in News feeds

#### 6.2.1. Data Collection

We collected data from 96 news feeds[5] during the year of 2011. Bag-of-Word (BoW) features were extracted using standard natural language processing tools[6]. After removal of stop words and stemming our BoW dictionary contained $W \approx 10^5$ words. The time series of each word was tf-idf normalized. The feature time series were then stored in sparse matrices $X_f \in \mathbb{R}^{W \times T}$ where $f = \{1,\ldots,F = 96\}$ denotes news feed and $t = \{1,\ldots,T\}$ denotes the time in hours. Time stamps of all news web sources were set to CET. For the sake of comprehensibility in the results presented here we focus on the month of October in 2011. In this month a clearly detectable trend were reports of Steve Jobs' death.

---
[5]http://beta.wunderfacts.com/
[6]http://www.nltk.org/

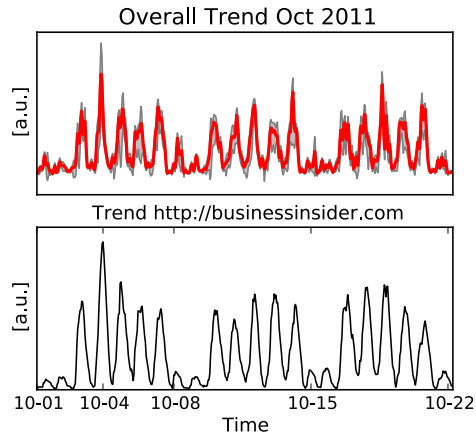

Figure 2. *Top panel:* Canonical trends $y_f(t)$ in arbitrary units (*a.u.*, each trend $y_f(t)$ was normalized to $\sum_t y_f(t)^2 = 1$) during the first three weeks of October 2011 (median over all feeds in red and 25th/75th percentiles in gray). Note the weekly oscillations, 5 peaks for each working day and a trough for the weekends. Steve Jobs' death marks a strong peak in the first week of October. *Bottom panel:* The prediction $\hat{y}_f(t)$ obtained from the news feed http://businessinsider.com showed the highest prediction accuracy of this trend.

#### 6.2.2. Canonical Trends in News feeds

As we obtain a canonical projection $w_y$ for each pool of web sources $Y_f$, the canonical trends that are predicted by each news feed $X_f$ could potentially differ. In practice however, the canonical trends are very similar. Figure 2 shows in the top panel the median and 25th/75th percentiles of all canonical trends in October 2011. The percentiles are very close to the median trend, indicating a large similarity of the different canonical trends. Reports on Steve Jobs' death mark a pronounced peak in the first week reflected in all canonical trends. Also note that the trends clearly reflect the weekly publishing activity on the news feeds, five peaks each week and a trough reflecting the weekend. Our results show that the temporal dynamics in the canonical subspace can be easily interpreted and authentically reflect the impact of relevant information cascades in large web graphs.

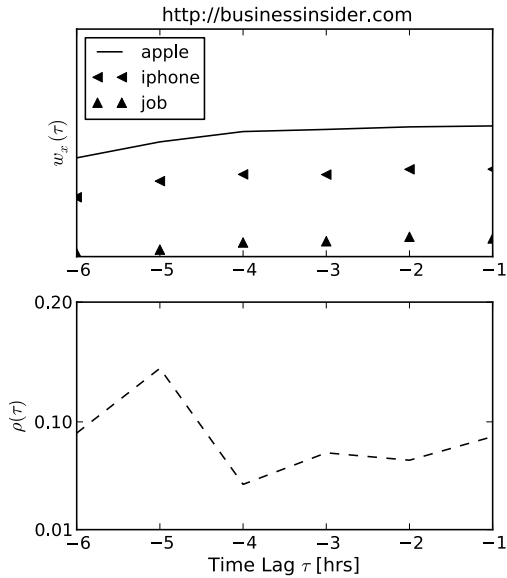

| Website | 25th/50th/95th Percentile CT | Median $CT_{\text{shuffled}}$ | LSA |
|---|---|---|---|
| businessinsider | 0.74/0.80/0.81 | 0.05 | 0.74 |
| arstechnica | 0.54/0.67/0.72 | 0.03 | 0.60 |
| engadget | 0.60/0.67/0.73 | 0.04 | 0.67 |
| techcrunch | 0.47/0.64/0.70 | 0.13 | 0.63 |
| mashable | 0.53/0.63/0.70 | 0.08 | 0.57 |
| venturebeat | 0.57/0.62/0.70 | 0.09 | 0.52 |
| techdirt | 0.39/0.61/0.70 | 0.08 | 0.55 |
| theregister | 0.47/0.56/0.67 | 0.15 | 0.58 |
| forbes | 0.48/0.55/0.69 | 0.07 | 0.47 |
| guardian | 0.47/0.53/0.58 | 0.08 | 0.56 |

*Table 1.* Top ten trend setter news feeds and their prediction accuracies, normalized as canonical correlation. Left column shows results of *Canonical Trend* (CT) algorithm, middle column results of CT on shuffled data ($CT_{\text{shuffled}}$) and right column the predictions obtained by LSA on $X_f$ and $Y_f$ separately (see sec 5.4).

*Figure 3. Top panel:* Canonical convolution $w_x(\tau)$ corresponding to the best predicting news feed http://businessinsider.com (see fig. 2); plotted are normalized weights of the top three words. Words constituting the trend in fig. 2, *bottom*, were associated with Steve Jobs or Apple. *Bottom panel:* The canonical correlogram $\rho(\tau)$ (see eq. 6) for the top predictor reflects the temporal dynamics between the single feed and all others. A peak at $\tau = -5$hrs indicates an increased prediction accuracy of all other feeds five hours in advance.

6.2.3. Canonical Trend Prediction

We investigated how well we can predict the trends in a pool of web sources from a single web source. Table 1 shows the prediction accuracy as canonical correlation (see eq. 5) for the ten best predictors, i.e. the trend setter news feeds, summarized as 25th/50th/75th percentiles across cross validation folds. Using the information published at $t - \tau$, $\tau = \{1, \ldots, 5\}$, meaning five to one hours before all other feeds, the listed news feeds could predict the overall trend at time $t$ with high accuracy. For instance the web site http://businessinsider.com predicted the content of all other news websites in the data set in more than 50% of the cases tested with a correlation coefficient of 0.8. The trend prediction $\hat{y}_f(t)$ of the top trendsetter http://businessinsider.com is shown in the bottom panel of figure 2. The time course clearly captures the temporal variation of the overall trend, depicted above in the top panel of fig. 2. In the top panel of figure 3 the time lag dependent features of $w_x(\tau)$ are depicted. The words to which the canonical trend detection algorithm assigned high weights were associated with Steve Jobs or Apple. The corresponding canonical correlogram $\rho(\tau)$ has a pronounced peak at $\tau = -5$hrs.

It is important to note that the temporal dynamics of single features in $w_x(\tau)$ can be different than those of $\rho(\tau)$. One reason for this is that $w_x(\tau)$ is non-separable, meaning that it does not factorize into a single temporal component and one component that describes the dependencies in the BoW feature space. So in order to get the full picture of the temporal dynamics between $X_f$ and $Y_f$ one has to look at the time courses of all features in $w_x(\tau)$. However we can identify relevant features from $w_x(\tau)$ by picking those with the highest absolute weights, summed over time lags. And we can extract the overall temporal dynamics from the canonical subspace from $\rho(\tau)$.

We compared the predictions from the canonical trend algorithm to predictions obtained with a standard topic detection method (LSA, see section 5.4). The results are shown in table 1, *right column*. In the LSA setting, we extracted topics from the BoW time series of single news feeds and the average BoW time series separately. Predictions of the strongest topics in all news feeds based on the strongest topics in a single feed are lower than the CT predictions. This suggests that canonical trends found in a single web source generalize better to a pool of web sources. This is expected from the different objective functions of CT (see eq. 5) and LSA (see eq. 11).

## 7. Conclusion and Outlook

We presented a simple, efficient and purely data driven method for detecting news trends and trend setters in web data. By making use of the kernel trick we can efficiently exploit the full multivariate structure of temporal dependencies in the canonical subspace of web graph features such as the BoW representation. Both the detected trends and the features learned by the algorithm authentically reflect the true impact of information cascades in temporally evolving graphs. Future work includes more empirical evaluations to study temporal correlation not only from BoW features but also from auxiliary data, such as frequency of retweets along the lines of (Lerman & Hogg, 2010), which predict popularity of content based on early user interest. Another useful feature representation could be named entities along the lines of (Gabrilovich et al., 2004). Independent of the feature representation employed it is important to note that the CT algorithm is unsupervised. The objective of the CT algorithm, maximal co-variation (eq. 5), does not necessarily yield the most interesting trends. Some information that is highly relevant might not be reflected as the main oscillation in all news feeds. However the criterion used in our approach, maximal variance explained, is useful if one is interested in the web sources that have the strongest overall impact. For more detailed analyses the trend of interest could be manually defined (for instance by picking only a few words of interest). Future research will also have to investigate temporal interactions between multidimensional canonical trend subspaces. Moreover we here assumed that the temporal dependencies between web sources are stationary in the analysis period. In general this might not be the case. Web source dependencies can be highly non-stationary. These non-stationarities have to be investigated using appropriate methods, as for instance (von Bünau et al., 2009).

### Acknowledgements

We thank Klaus-Robert Müller and Manuel Gomez-Rodriguez for helpful discussions. M.B. and F.B. would like to acknowledge support for this project from the BMBF project ALICE, "Autonomous Learning in Complex Environments" (01IB10003B).